# Move from Perturbed scheme to exponential weighting average

Chunyang Xiao

Abstract

In an online decision problem, one makes decisions often with a pool of decisions' sequence called experts but without knowledge of the future. After each step, one pays a cost based on the decision and observed rate. One reasonal goal would be to perform as well as the best expert in the pool. The modern and well-known way to attain this goal is the algorithm of exponential weighting. However, recently, another algorithm called follow the perturbed leader is developed and achieved about the same performance. In our work, we first show the properties shared in common by the two algorithms which explain the similarities on the performance. Next we will show that for a specific perturbation, the two algorithms are identical. Finally, we show with some examples that follow-the-leader style algorithms extend naturally to a large class of structured online problems for which the exponential algorithms are inefficient.

## 1 Online problem setting

In an online decision problem, one makes decisions often with a pool of decisions' sequence called experts but without knowledge of the future. After each step, one pays a cost based on the decision and observed state. As there is no prior knowledge on the accuracy of experts in the pool, one reasonable goal for this general problem would be to perform as well as the best expert in the pool after a number of steps. More precisely, we consider the following mathematical problem:

- A set S of experts is given.

- The algorithm interacts with an adversary in a series of T steps.

- In each step j, the algorithm picks an expert $x_j \in S$, and the adversary selects a cost function $c_j$: S → R. The adversary could be adaptive, in that $c_j$ may depend on $\{x_i : i < j\}$.

- The algorithm incurs cost , and receives as feedback the value of $c_j(x_j)$.

- Minimize the algorithm's regret which is defined as difference in expected cost between the algorithm's sequence of choices and that of best fixed expert in S:

$$Regret = E\left[\sum_j c_j(x_j)\right] - min_{x \in S} \sum_j E[c_j(x_j)]$$

This version formulates the problem typically as predicting from expert advice problem. We consider here that we could only choose from the decision space of the experts. And we consider here only the full transparent model, *i.e.* after each step, all

One immediate consequence of this hypothesis is that for general weighted average forecaster, the given weighted average may not be in the decision space. However, we could get around this problem by reinterpreting the weights not as real weights but the probabilities in decision space. This is little bit more complicated than working in a continuous space but this framework will allow us to compare general weighted average forecaster with other forecasters perturbed leading expert more easily. Besides that, it occurs often that the decision space happens to be all the experts' decisions, *i.e.* at practice, we have to choose one expert to listen to.

Another remark that the problem that we describe here is very general and of course could include many specific kinds of problems. For example, in shortest paths problems, $c_j$ are linear functions [1]. For certain algorithms (follow perturbed leading expert, for example), the functions $\min_{x \in S} \sum_j E[c_j(x_j)]_j$ would be needed to be calculated efficiently. This is really not a strict condition, as it satisfies for at least all convex functions.

## 2    Two approaches

The modern and well known way to tackle the problem is exponential weighting scheme that have been discovered and rediscovered in many areas. A survey of these results could be found in [3]. However, in recent years, another scheme called follow perturbed leading expert has been discovered and achieved similar performance [2]. In this section, we will detail the two different approaches with some variants of the two approaches. Some theoretical bounds which have already been proven would also be indicated in this section. We remark in particular some similarities in theoretical bounds for two approaches which suggest some relationship between the two approaches. This relationship will be studied further in the next section.

As its name suggests, weighted average forecaster forecasts the future based on a weighted average of the prediction of experts. Each expert is associated with a weight that we assign with its historic performance (the weight could be naturally updated after each step). After that, we take our decision as the weighted decision of experts. When the space is convex, it is possible for the forecaster to give rise to a new predicted decision value, different from decisions coming from all experts. Otherwise, in the case that we consider here, the weight would be considered as probability to take certain precisions. For example, in algorithm 1, the term $\frac{e^{-\beta L_{i,t-1}}}{\sum_{i=1}^{n} e^{-\beta L_{i,t-1}}}$ is the probability (weight) to take the decision of expert i at step t.

The exponential weighted average forecaster we consider is the following:

**Algorithm 1 weighted average forecaster**

At each t, calculate loss of each expert amongst n experts:

$L_{i,t-1} = \sum_{j=1}^{t-1} c_{i,j}(f_{i,j})$, $f_{i,j}$ is the choice of expert i at step j

Choose the decision of expert i at t with probability:

$$\frac{e^{-\beta L_i, t-1}}{\sum_{i=1}^{n} e^{-\beta L_i, t-1}}$$

---

First let's remark that the weighted average is a generalization of the randomized weighted majority algorithm that appeared in the survey of Blum [3]. The problem considered there is more strict, the loss would be specified only by 0 or 1. In other words, there is only one right answer and one wrong answer. At each step, randomized weighted majority algorithm updates the weight of wrong answer by multiplying its weight by a constant $\gamma \leq 1$. By taking $\gamma = e^{-\beta}$ it could be easily seen that weighted majority is just a special case of weighted average forecaster that we consider here. For randomized weighted majority algorithm, a bound is proven with the following theorem:

*Theorem 1 On any sequence of trials, the expected loss noted M made by randomized weighted majority algorithm satisfies:*

$$M \leq \frac{m \ln\left(\frac{1}{\beta}\right) + \ln n}{1 - \beta}$$

*where n is the number of experts and m the minimal loss of all the experts so far.*

Now, we will consider an alternative approach to tackle the problem. Instead of taking a weighted average of all the experts, the most intuitive way would be to follow the best leading expert. However, this very intuitive idea does not perform well on some very simple cases. Consider a two-expert system that one always predicts 1 and the other always predicts 0. The cost sequence is (0,0.5) followed by alternating (1,0) and (0,1). Then, after n step, following the leader would lead to a loss of n well the best expert generates a loss of only n/2. Remark that weighted average performs well in this special case (as well as theory tells) because it leaves a possibility to the expert that performs a little less well in the history and thus avoid the tragedy of choosing always the bad expert. Another way to leave this possibility to all the experts and to not fix one is to introduce a perturbation before making the decision. This idea leads us to the following perturbed leading expert algorithm presented in the paper of Adam Kalai [2].

---
**Algorithm 2** follow the perturbed leading expert
---

On each step t = 1, 2, …

1. For each expert $e \in \{1,2,…\}$, pick $p_t[e] \geq 0$ from a distribution.

2. Choose expert with minimal $c_t[e] - p_t[e]$, where $c_t[e]$ = total cost of expert e so far.

---

Remark first that we have a great amount of liberty here in this algorithm. The distribution is free to our choice and indeed, different choices of distributions induce different bounds. And according to situations, some distributions may be preferred comparing to others. Secondly, if this algorithm obtains about the same performance as weighted average algorithm, it would be still more interesting than the average weighted version. Observe that what we need here to choose the minimum is only a minimization oracle [1] as we do not update weight and thus modify the structure of the problem. Put it into another way, we have decoupled the problem with on the one hand, the search of perturbation, and on the other hand, a deterministic optimization problem which has been elegantly solved in various domains. Thus, our algorithm leads generally to a gain in speed. This point is particularly illustrated in the last section of applications with perturbation schemes.

To show the bound results of this class of algorithms. Let note D the decision domain and S the cost domain, $|x|_1$ as norm $L^1$ of x. The following theorem is presented in [2] for the case when the cost functions are all linear. It consists of a theorem which gives the bound for uniform and exponential distribution (noted FPL and FPL* respectively):

$D \geq |d - d'|$, for all $d, d' \in D$
$R \geq |d.s|$, for all $d \in D, s \in S$
$A \geq |s|_1$, for all $s \in S$

*Theorem 2 Let $s_1, s_2, …, s_t$ be a cost sequence.*

*(a) Running FPL with parameter $\epsilon \leq 1$ gives*
$$E\left[FPL(\varepsilon)\right] \leq mincost_T + \epsilon RAT + \frac{D}{\varepsilon}$$

*(b) For non-negative $D, S \in R_+^n$, FPL\* gives*
$$E\left[FPL*(\frac{\varepsilon}{2A})\right] \leq (1+\varepsilon)mincost_T + \frac{4AD(1+lnn)}{\varepsilon}$$

Remark that in particular, the bound given by FPL* resembles a lot the bound for randomized weighted majority given in theorem 1. We will show in the next section that indeed there exists a strong relationship between these two approaches. This relationship would be not only important in theoretical point of view but also important in practice. As the penalty of weights offers more insight on what kind of penalization we are applying whereas the follow perturbed leading expert could be likely much faster in running time. We will come up to this point later on.

## 3  Relationship between two approaches

In this section, we show the properties which are shared by both FPL* and weighted average forecaster shown in algorithm 1. Recall the simple fact that weighted average forecaster is a generalization of weighted majority algorithm. After that, we apply the same techniques to deduce a "weighted average forecaster" for uniform distribution. The calculation to process the conversion will be shown when we deduce the first example.

First consider the weighted average forecaster. At each step t, the choice of our algorithm is totally characterized by the probability (weight) attributed to the decision of each expert. Thus, it is also totally characterized by the probability ratio between any two probability (weight) . According to weighted average forecaster algorithm, at step t, this ratio could be written as:

$$\frac{P(expert_i)}{P(expert_j)} = e^{-\beta(L_{i,t-1} - L_{j,t-1})}$$

This ratio has an intuitive explanation. When $L_{i,t-1} - L_{j,t-1} \leq 0$, expert i is chosen with a probability much larger that expert j. That means we prefer the expert that generated small loss historically yet leave always a possibility to choose an expert which performed less well in the past. This possibility move towards one when the difference becomes negligible. Seen in the formula, the probability ratio is quantified as the exponential of the loss difference.

Now let's examine the case of algorithm follow the perturbed leading expert. Intuitively, as a perturbation is added to each of the experts, we know that a possibility is generated to choose each expert. It is question to appropriately choose the perturbation distribution so that the previous probability ratio found in weighted average forecaster remains. In the next, we will show that the exponential distribution is a distribution that generates very similar conditions compared to the above condition.

Note c the loss difference between expert i and expert j. Thus, $= L_{i,t-1} - L_{j,t-1}$. Without loss of generality, we suppose here that $c \geq 0$. Let us calculate the probability that the algorithm choose expert i over the other expert j.

$$\frac{P(expert_i)}{P(expert_i) + P(expert_j)} = P(c + d_i \leq d_j) \tag{1}$$

where $d_i$ and $d_j$ are instances generated by independent exponential distributions. For the inequality holds, we consider for example $d_i$ take a value v and $d_j$ take a value more than v + c . As $d_i$ and $d_j$ are independent, the joint probability of the two events are written simply as the multiplication of the two terms. Thus, the above probability is the sum over all the possible v when the distribution μ has probability mass functions (particularly, it takes a finite number of values):

$$P(c + d_i \leq d_j) = \sum_v P_\mu(v) P_\mu(x \geq v + c)$$

In the case that the distribution has probability density functions (particularly, it takes an infinite number of values), the sum becomes an integral:

$$P(c + d_i \leq d_j) = \int f_\mu(v) \left( \int f_\mu)(x)\, dx \right) dv, x \geq v + c$$

Now take the exponential distribution into the above equation (2) , we have:

$$P(c + d_i \leq d_j) = \int_0^{+\infty} \epsilon e^{-x\epsilon} \left( \int_{x+c}^{+\infty} \epsilon e^{-y\epsilon}\, dy \right) dx$$

$$= \int_0^{+\infty} \epsilon e^{-y\epsilon} . \epsilon e^{-(x+c)\epsilon} dx = \frac{1}{2} e^{-\epsilon x}$$

Let us put weighted average forecaster probability that we take now in the same form:

$$\frac{P(expert_i)}{P(expert_i) + P(expert_j)} = \frac{e^{-\beta c}}{1 + e^{-\beta c}}$$

The two formulas are quite similar in the way that:

- When the c is near zero, the two ratio gives unbiased probability 0.5
- When the c is large, the two ratio gives an exponential dependence, the algorithm follow the perturbed leader add a factor 0.5 compared to weighted average forecaster.

For another example, let us consider the algorithm follow the perturbed leader with uniform distribution in $[0,\epsilon]$. In this case, it makes sense to suppose that $c \leq \epsilon$ as otherwise, the probability of choose expert i over expert j would be 0. We suppose $c \geq 0$ as usual, then by following the same calculus by using be 0. We suppose equation (1) and (2), we have:

$$\frac{P(expert_i)}{P(expert_i) + P(expert_j)} = P(c + d_i \leq d_j)$$

$$= \int_0^{\epsilon-c} \frac{1}{\epsilon} \left( \int_{x+c}^{\epsilon} \frac{1}{\epsilon} dy \right) dx = 1 - \frac{1}{2\epsilon^2}(\epsilon^2 - c^2 + 2\epsilon c)$$

We could easily verify that c = 0 corresponds to a probability ratio 0.5, and $c = \epsilon$ corresponds to a probability ratio 0. It means that the model becomes deterministic from a certain threshold $\epsilon$ in contrary to the weighted average forecasting and the follow exponential perturbed leading expert algorithm. Meanwhile, we could note that the penalization is in the quadratic form. Thus compared to the previous perturbation, we have less penalization in loss here and lose probability or certain experts from a threshold. These two points are supposed to explain why this model possesses a bound that still depends on the step T.

## 4   Towards exponential weighting

The distribution that we consider here is Gumbel distribution. Before the calculus, we will give a brief description about gumbel distribution which inspires us to consider this particular distribution. Gumbel distribution is frequently used to model the distribution of the maximum (or the minimum) of a number of samples of various distributions, thus such a distribution might be used to represent the distribution of the maximum level of a river in a particular year if there was a list of maximum values for the past ten years. It is a particular case of the generalized extreme value distribution. The very property that leads us to consider this distribution is stated as the following:

*Property The difference of two Gumbel-distributed random variables has a logistic distribution.*

This gives birth directly to the perturbation result on probability ratio:

$$\frac{P(expert_i)}{P(expert_i) + P(expert_j)} = P(d_j - d_i \geq c) = \frac{e^{-\frac{c}{\beta}}}{1 + e^{-\frac{c}{\beta}}}$$

As in the previous section, c is the loss difference between expert i and expert j realized until considered step. It is obvious that by replacing β here by $\frac{1}{\beta'}$, we obain the weighting average result. Thus we announce the following theorem:

*Theorem 3 The follow perturbed leading expert algorithm with perturbing distribution as gumbel distribution $e^{-z-e^{-(x-\mu)/\beta}}$ is equivalent to an weighted majority algorithm with $\beta' = \frac{1}{\beta}$.*

## 5 Application of perturbation schemes

These follow-the-leader style algorithms extend naturally to a large class of structured online problems for which the exponential weighted average algorithms are inefficient[2]. This section illustrates some of these examples.

### 5.1 Online shortest path problem

In this problem, one has a directed graph and a fixed pair of nodes (s, t). Each period, one has to pick a path from s to t, and then the times on all the edges are revealed. The per-period cost is the sum of the times on the edges of the chosen path. This problem could be viewed as an expert problem where each path consists of an expert. Naive weighted average algorithm could be applied to this problem but has high complexity as each (expert) path should be considered and their weights should be updated. Some clever schemes have been developed[5], but now let us consider just the algorithm Follow the perturbed leading expert:

On each period t = 1, 2,...,
    1.      For each edge e, pick $p_t[e]$ randomly from an gumbel distribution.
    2.      Use the shortest path in the graph with weights $c_t[e] + p_t[e]$ on edge e, where $c_t[e]$ = total time on edge e so far.

It could be seen that this algorithm does not involve any update weight calculation. As no weight update is used, what we need, as mentioned in[1], is just an optimization oracle that we have here Bellman algorithm, for example. And this algorithm achieves similar bounds as in [5].

## 5.2   Image segmentation

Image segmentation is a middle level image processing problem. One very popular approach to tackle this problem is to use MRF (Markov Random Fields). While there are many researches on the deterministic minimization on these fields and achieve good results, the accurate probability inference behind these problems generally require very time consuming MCMC techniques. The article [4] shows that with the addition of perturbation, the probability inference could be carried on easily. Besides, these techniques could be extended to parameter estimation by using moment matching rule.

# 6   Conclusion

We show that for expert advice problem, the follow perturbed leading expert algorithm could be equivalent to the modern exponential weighted average algorithm by carefully choosing the perturbation distribution. We show that different perturbation distribution may be chosen according to different circumstances and the distribution choice could be interpreted as the penalization on the weights when we interpret the perturbed leading expert algorithm as a weighted average algorithm. We further argue that the separation of the online optimization problem into its online and offline components which appears in follow perturbed leading expert algorithm is helpful as only deterministic optimization oracle is needed for offline components.

# References

[1] Baruch Awerbuch and Robert D Kleinberg. Adaptive Routing with Endto-End feedback : Distributed Learning and Geometric Approaches. (x), 2004.

[2] Adam Kalai and Santosh Vempala. Efficient algorithms for online decision problems. *Journal of Computer and System Sciences*, 71(3):291–307, October 2005.

[3] Avrim L.Blum. On-line algorithms in machine learning. 1997.

[4] George Papandreou and Alan L. Yuille. Perturb-and-MAP random fields: Using discrete optimization to learn and sample from energy models. *2011 International Conference on Computer Vision*, pages 193–200, November 2011.

[5] Eiji Takimoto and Manfred K. Warmuth. Predicting nearly as well as the best pruning of a planar decision graph. *Theoretical Computer Science*, 288(2):217–235, October 2002.